\ifdefined\pdfoutput\pdfoutput=1\fi %
\documentclass{article}

\usepackage{style_iclr2027/fancyhdr}
\usepackage{style_iclr2027/iclr2027_conference,times}

\usepackage{url}            %
\usepackage{booktabs}       %
\usepackage{amsfonts}       %
\usepackage{fix-cm}         %
\usepackage{nicefrac}       %
\usepackage{microtype}      %
\usepackage{xcolor}         %

\usepackage{iftex}
\ifPDFTeX
  \usepackage[utf8]{inputenc}
  \usepackage[T1]{fontenc}
  \usepackage{CJKutf8}
  \AtBeginDocument{\begin{CJK*}{UTF8}{gbsn}}
  \AtEndDocument{\end{CJK*}}
\else
  \usepackage{fontspec}
  \usepackage{xeCJK}
  \defaultCJKfontfeatures{}
  \IfFontExistsTF{Songti SC}{\setCJKmainfont{Songti SC}\setCJKmonofont{Songti SC}}{\setCJKmainfont{FandolSong-Regular.otf}\setCJKmonofont{FandolSong-Regular.otf}}
\fi
\usepackage{amsmath,amssymb,graphicx,array,tabularx,needspace,caption,subcaption}
\usepackage{wrapfig,colortbl}
\definecolor{defaultrow}{gray}{0.94}
\usepackage[section]{placeins}
\providecommand{\tightlist}{\setlength{\itemsep}{0pt}\setlength{\parskip}{0pt}}
\usepackage[pagebackref,breaklinks,colorlinks]{hyperref}

\usepackage{cleveref}
\usepackage{xspace}
\usepackage{lipsum}

\usepackage{algorithm}
\usepackage{listings}
\definecolor{codecomment}{rgb}{0.25,0.5,0.5}
\definecolor{codekw}{rgb}{0.63,0.11,0.51}
\usepackage{enumitem}
\definecolor{citegreen}{rgb}{0,0.5,0}
\hypersetup{citecolor=citegreen}

\newcommand{\gr}[1]{{\hypersetup{citecolor=black!45}\color{black!45}#1}}
\definecolor{oursrow}{rgb}{0.91,0.95,1.0}

\title{Embedding Prediction Helps Image Generation}
\definecolor{projecturl}{rgb}{0.2,0.45,0.75}

\author{Sihan Xu$^{1}$\thanks{Equal contribution.} \quad Ji Xie$^{2}$\footnotemark[1] \quad Zilin Wang$^{1}$ \quad Hui Shen$^{1}$ \quad Stella X. Yu$^{1}$ \\
$^{1}$University of Michigan \quad $^{2}$Carnegie Mellon University \\
{\hypersetup{urlcolor=projecturl}\url{https://sihanxu.me/nepa-dit}}}

\iclrfinalcopy %

\begin{document}

\maketitle
\lhead{Preprint} %
\hypersetup{
  pdftitle={Embedding Prediction Helps Image Generation},
  pdfauthor={Sihan Xu, Ji Xie, Zilin Wang, Hui Shen, Stella X. Yu},
  pdfkeywords={image generation, diffusion transformers, next-embedding prediction, multi-embedding prediction}
}
\raggedbottom
\renewcommand{\topfraction}{0.9}
\renewcommand{\bottomfraction}{0.8}
\renewcommand{\textfraction}{0.1}
\renewcommand{\floatpagefraction}{0.75}
\setcounter{topnumber}{4}
\setcounter{bottomnumber}{3}

\begin{abstract}
In diffusion transformers, a class label or a text prompt is embedded once, and the same condition is reused at every denoising step. We ask whether predicted embeddings can serve as this condition instead. Next-Embedding Predictive Autoregression (NEPA) trains a Transformer to predict the next continuous embedding in a sequence. In generation, the clean image follows the noisy image, so its embeddings are the next embeddings after the condition and the noisy image. We train a NEPA model to predict them all at once with \textbf{Multi-Embedding Prediction}, and in \textbf{Embedding Conditioned Generation}, a DiT generator is conditioned on these predictions, recomputed at every denoising step, so the conditioning signal adapts to the current noisy state. Experiments on class-conditional ImageNet \(256\times256\) study the condition of the generator, the design of Multi-Embedding Prediction, and the scaling of both models. The NEPA model adds a second network to every sampling step; with it, and combined with REPA, our final model, \textbf{NEPA-DiT-XL}, reaches an FID of \textbf{1.32} using about a third of the training compute of REPA.
\end{abstract}

\Needspace*{4\baselineskip}
\section{Introduction}
\begin{wrapfigure}{r}{0.42\textwidth}
\centering
\captionsetup{justification=raggedright,singlelinecheck=false}
\includegraphics[width=\linewidth]{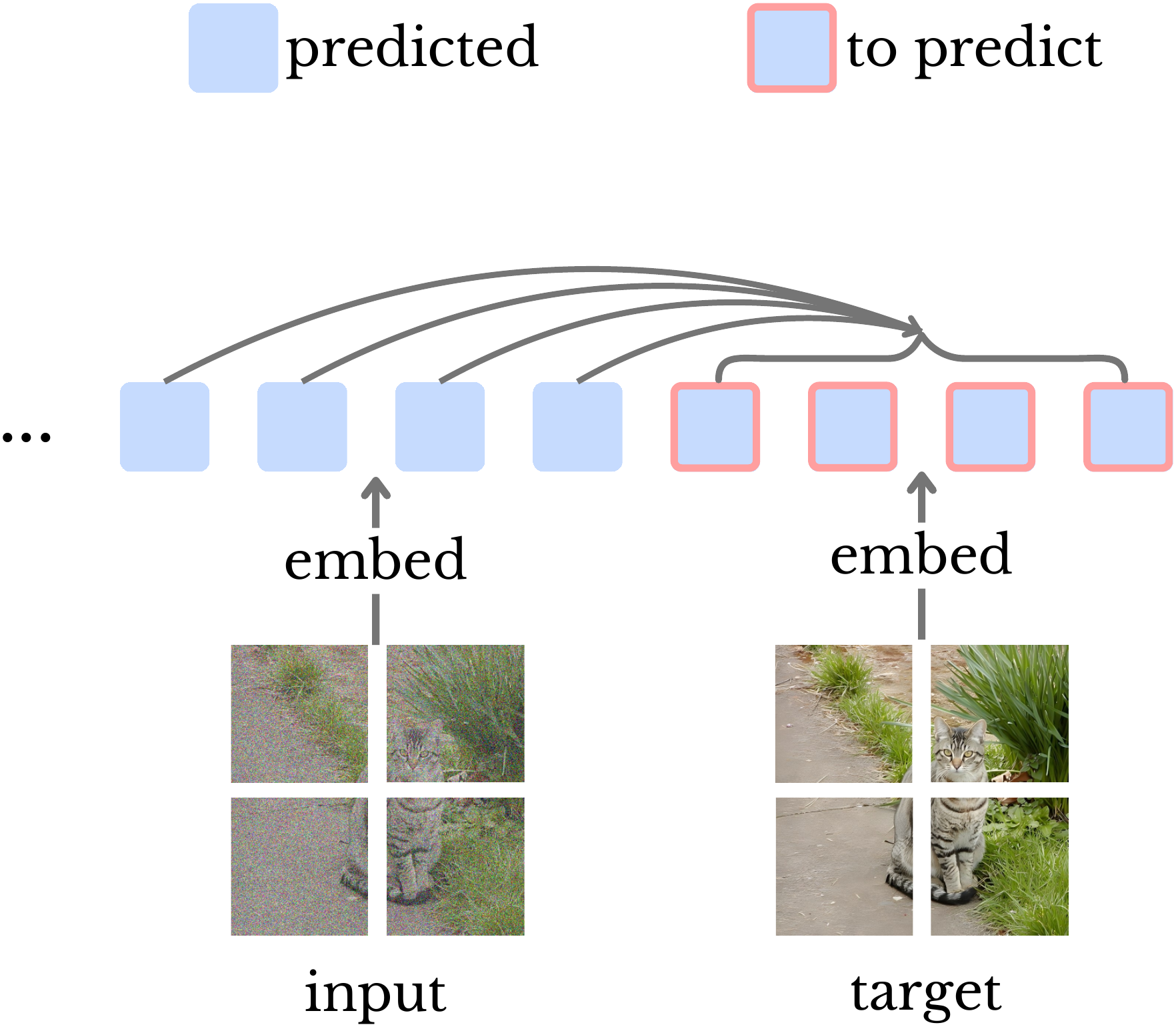}
\caption{\textbf{Multi-embedding prediction.} From the condition and the noisy image, the model predicts the next embeddings in the sequence, those of the clean image. Supervision is applied in embedding space; the predicted embeddings then condition a diffusion generator.}
\label{fig:mep}
\vspace{-10pt}
\end{wrapfigure}

Diffusion transformers have become the standard backbone for image generation~\citep{peebles2022dit,ma2024sit,esser2024sd3}. Across these models, the condition enters through a fixed interface: a class label or a text prompt is embedded once, and the same condition is reused at every denoising step. It is left to the generator to work out what the condition implies for the image at hand as its state moves from noise to a clean sample. Recent work brings pretrained visual representations into this pipeline, as alignment targets for the generator's features~\citep{yu2024repa}, as the latent space in which diffusion operates~\citep{zheng2025rae}, or as intermediate generation targets~\citep{li2023rcg}.

In all of these, the condition itself stays fixed while the image evolves. We ask: \emph{can the condition itself be predicted from the current noisy image, at every denoising step?}

NEPA trains a Transformer to predict the next embedding in a sequence, rather than to reconstruct pixels or discrete tokens~\citep{xu2025nepa}. Unlike a feature extractor, such a model is predictive by construction: its outputs are estimates of embeddings it has not yet seen. In generation, the sequence runs from the condition through the noisy image to the clean image, so the clean image is what comes next, and its predicted embeddings are a natural candidate for what a generator should condition on.

We answer this question with \textbf{Embedding Conditioned Generation} (ECG): a DiT \emph{generator} is conditioned on the embeddings predicted by a frozen NEPA model instead of on the raw condition, and the NEPA model is queried at every denoising step, so the conditioning signal adapts to the current noisy state. The NEPA model is trained with \textbf{Multi-Embedding Prediction} (MEP), which extends NEPA to predict the next several embeddings at once, as multi-token prediction does for tokens in language modeling~\citep{gloeckle2024multitoken}; it reads the condition and the noisy image and predicts the embeddings of the clean image (Figure~\ref{fig:mep}).

On class-conditional ImageNet \(256\times256\), we compare NEPA predictions with networks of the same size trained in other ways, and study the design of Multi-Embedding Prediction and the scaling of both models. Combined with REPA, our final model, NEPA-DiT-XL, reaches an FID of \textbf{1.32}.

We highlight the main contributions of this paper below:
\begin{itemize}[leftmargin=*]\tightlist
\item We propose Embedding Conditioned Generation, which conditions a DiT generator on the clean embeddings predicted by a NEPA model from the condition and the noisy image, recomputed at every denoising step.
\item We train the NEPA model with Multi-Embedding Prediction, which extends NEPA to predict the embeddings of the whole clean image at once.
\item On class-conditional ImageNet \(256\times256\), we study the condition of the generator, the design of Multi-Embedding Prediction, and the scaling of both models; combined with REPA, NEPA-DiT-XL reaches an FID of 1.32.
\end{itemize}

\Needspace*{4\baselineskip}
\section{Related Work}
\Needspace*{4\baselineskip}
\subsection{Diffusion Models and Representations for Generation}\label{sec:related-diffusion}

Denoising diffusion models have become a dominant framework for image generation~\citep{song2019score,ho2020ddpm,song2020score_sde,nichol2021improved,dhariwal2021adm}, supported by faster samplers~\citep{song2020ddim,lu2022dpmsolver,song2023consistency} and by flow-based formulations~\citep{lipman2022flowmatching,liu2022rectifiedflow,ma2024sit,huang2026recfm}. Latent diffusion models operate in the latent space of a pretrained autoencoder to reduce the computational cost~\citep{rombach2021ldm,esser2020vqgan,chen2024dcae,yao2025lightningdit}, while diffusion transformers replace U-Net denoisers with more scalable architectures~\citep{peebles2022dit,bao2022uvit}. Text-to-image models further improve these systems through stronger architectures, flow-based objectives, and text conditioning~\citep{nichol2021glide,ramesh2022dalle2,saharia2022imagen,balaji2022ediffi,podell2023sdxl,chen2023pixart,chen2024pixartsigma,esser2024sd3,li2024hunyuan,gao2024lumina,xie2024sana,xie2026reca}. Beyond synthesis, diffusion models have been adapted to text-guided image manipulation and editing~\citep{meng2022sdedit,hertz2023prompt2prompt,mokady2023nulltext,brooks2023instructpix2pix,xu2023cyclenet,xu2024infedit}, to spatially controlled generation~\citep{zhang2023controlnet}, and to temporally consistent video generation~\citep{ho2022videodiffusion,guo2024animatediff,xia2024unictrl}. Pretrained visual representations have been brought into this pipeline as generation targets~\citep{li2023rcg,wu2025reg,kouzelis2025redi}, alignment signals for diffusion features~\citep{yu2024repa,wang2025ddt,jiang2025sra,tian2025urepa,leng2025repae}, and the generative latent space~\citep{zheng2025rae,tong2026scaling_rae,singh2026raev2}. In these models, the class or text condition is injected by adaLN~\citep{peebles2022dit} or joint attention~\citep{esser2024sd3} and stays fixed across denoising steps. Unified multimodal models condition a diffusion decoder on embeddings produced by a language model~\citep{sun2024emu,ge2024seed,pan2025metaqueries,chen2025blip3o}, but these embeddings are likewise computed once before sampling. Self-conditioning~\citep{chen2023analogbits} feeds the estimate of the clean sample from the previous sampling step back into the denoiser, and RIN~\citep{jabri2023rin} carries latent states across steps; both reuse the estimates of the denoiser itself. We condition the generator on the predictions of a separate NEPA model, trained with its own embedding prediction objective and recomputed from the current noisy input at every step.

\Needspace*{4\baselineskip}
\subsection{Predictive Representation Learning}\label{sec:related-predictive}

Self-supervised visual learning has progressed from handcrafted pretext tasks~\citep{doersch2015context,noroozi2016jigsaw,zhang2016colorization,gidaris2018rotation} to contrastive and self-distillation methods~\citep{oord2018cpc,he2019moco,chen2020simclr,grill2020byol,caron2020swav,caron2021dino,zhou2021ibot} and to masked reconstruction approaches that predict pixels, tokens, or visual features~\citep{bao2021beit,he2021mae,xie2021simmim,wei2021maskfeat,baevski2022data2vec}. Generative image models instead learn by predicting discrete or continuous visual tokens, either autoregressively or through masked and progressively refined prediction~\citep{oord2017vqvae,esser2020vqgan,yu2021vitvqgan,lee2022rqvae,chang2022maskgit,li2022mage,yu2024titok,tian2024var,sun2024llamagen,li2024mar}. However, pixel and discrete-token prediction can emphasize low-level details or tokenizer-specific structure. Predictive representation learning avoids direct reconstruction by predicting targets in a learned embedding space: JEPA and I-JEPA predict masked target representations from visible context, and V-JEPA extends this principle to spatiotemporal prediction~\citep{lecun2022jepa,assran2023ijepa,bardes2024vjepa,assran2025vjepa2}. NEPA further combines representation prediction with causal modeling by replacing the next-token prediction of language models~\citep{radford2019gpt2,brown2020gpt3} with next-embedding prediction~\citep{xu2025nepa}. Our Multi-Embedding Prediction extends NEPA to predict several embeddings at once, analogous to multi-token prediction in language modeling~\citep{gloeckle2024multitoken}.

\Needspace*{4\baselineskip}
\section{Method}
Our method has two stages. First, we train a NEPA model with \textbf{Multi-Embedding Prediction} to predict clean image embeddings from condition tokens and noisy image embeddings. Second, we train a DiT \emph{generator} with \textbf{Embedding Conditioned Generation}, where the frozen NEPA model produces predictive embeddings at each denoising step and the DiT uses them as its condition.

\Needspace*{4\baselineskip}
\subsection{Preliminaries}\label{sec:preliminaries}

\textbf{Next-embedding predictive autoregression.} NEPA replaces next-token prediction with next-embedding prediction~\citep{xu2025nepa}. An input
\(\mathbf{x}\) is split into \(N\) non-overlapping patches by a convolutional patch embedding
layer \(f\), giving a sequence of continuous embeddings \(\mathbf{z}=f(\mathbf{x})\). An autoregressive model \(h_\theta\) reads the sequence and predicts, at each position, the embedding that follows it, and is trained with a distance \(\mathcal{D}\) in embedding space between the predictions $\mathbf{p}$ and the targets $\mathbf{z}_{\text{target}}$:

\begin{equation}
\label{eq:nepa}
\mathbf{p} = h_\theta(\mathbf{z}),
\qquad\qquad
\mathcal{L}_{\mathrm{NEPA}} = \mathcal{D}(\mathbf{z}_{\text{target}}, \mathbf{p}),
\end{equation}

\noindent where \(\mathcal{D}\) is any distance between embeddings. For images, NEPA orders the patches into a sequence under causal attention, so the target is the next patch, \(\mathbf{z}_{\text{target}}=\mathbf{z}_{2:N}\).

\textbf{Latent diffusion and flow matching.} Following latent diffusion models and DiT~\citep{rombach2021ldm,peebles2022dit}, we perform image generation in the latent space of a pretrained autoencoder: a frozen encoder \(\mathcal{E}\) maps each image to a clean latent \(\mathbf{x}_0\), which serves as the input to both models, and a frozen decoder maps generated latents back to images. During training, we sample Gaussian noise \(\boldsymbol{\epsilon} \sim \mathcal{N}(\mathbf{0}, \mathbf{I})\) and a noise level \(t\), then construct the noisy input $\mathbf{x}_t = (1 - t) \mathbf{x}_0 + t \boldsymbol{\epsilon}$. We train the diffusion transformer to predict the flow target $\mathbf{v}_t = \boldsymbol{\epsilon} - \mathbf{x}_0$~\citep{ma2024sit,liu2022rectifiedflow}. Given a condition \(c\)\footnote{A class label for class-to-image generation, or a text prompt for text-to-image generation.}, the generator \(G_\phi\) is optimized with

\begin{equation}
\label{eq:flow-matching}
\mathcal{L}_{\mathrm{fm}}
= \mathbb{E}_{\mathbf{x}_0,\boldsymbol{\epsilon},t,c}
\Big[\,w(t)\,\big\|\,\mathbf{v}_t - G_\phi(\mathbf{x}_t, t, c)\,\big\|^2\Big],
\end{equation}

where \(w(t)\) is the timestep-dependent loss weighting. Conventionally, \(c\) is fixed across the sampling trajectory; we replace it with embeddings predicted by a NEPA model from both the condition \(c\) and the current noisy image state \(\mathbf{x}_t\). This allows the conditioning signal to adapt throughout denoising, as described in Section~\ref{sec:ecg}.

\Needspace*{4\baselineskip}
\subsection{Multi-Embedding Prediction}\label{sec:mep}

\begin{figure}[t]
\centering
\includegraphics[width=0.82\linewidth]{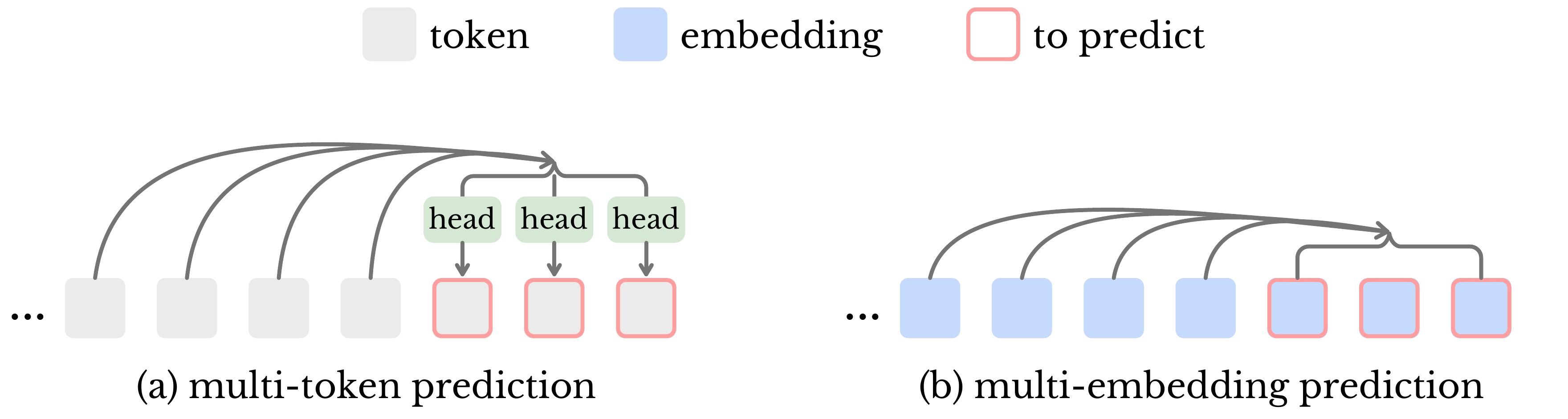}
\caption{\textbf{Multi-token vs.\ multi-embedding prediction.} Both predict several items at once. (a) Tokens are discrete, so hidden states are decoded into tokens by output heads. (b) Embeddings are continuous, so the outputs of the model are the predictions themselves. In both, the same context predicts all targets in one forward pass.}
\label{fig:mtp-vs-mep}
\end{figure}

NEPA predicts one next embedding at a time. In language modeling, predicting several future tokens at once improves over next-token prediction~\citep{gloeckle2024multitoken}. Predicting one embedding at a time is similarly limiting when the next embeddings form a group that is needed together, such as the patch embeddings of an image. \textbf{Multi-Embedding Prediction} (MEP) extends NEPA to predict the next \(K\) embeddings simultaneously from the same context. Given the context embeddings \(\mathbf{z}_{\le n}\), the model predicts the next \(K\) embeddings in a single forward pass:
\begin{equation}
\label{eq:mep-general}
\mathbf{p}_{n+1:n+K} = h_\theta(\mathbf{z}_{\le n}),
\qquad
\mathcal{L}_{\mathrm{MEP}} = \frac{1}{K}\sum_{k=1}^{K} \mathcal{D}(\mathbf{z}_{n+k}, \mathbf{p}_{n+k}).
\end{equation}
NEPA is the special case \(K=1\). Like multi-token prediction, MEP predicts several items at once; it differs in predicting in embedding space rather than token space. Predicted tokens must be decoded from hidden states by output heads, whereas predicted embeddings are the outputs of the model themselves (Figure~\ref{fig:mtp-vs-mep}). Section~\ref{sec:reasoner} instantiates MEP for image generation.

\Needspace*{4\baselineskip}
\subsection{NEPA for Generation}\label{sec:reasoner}

\vspace{4pt}
\noindent
\begin{minipage}[c]{0.32\textwidth}
\centering
\includegraphics[width=\linewidth]{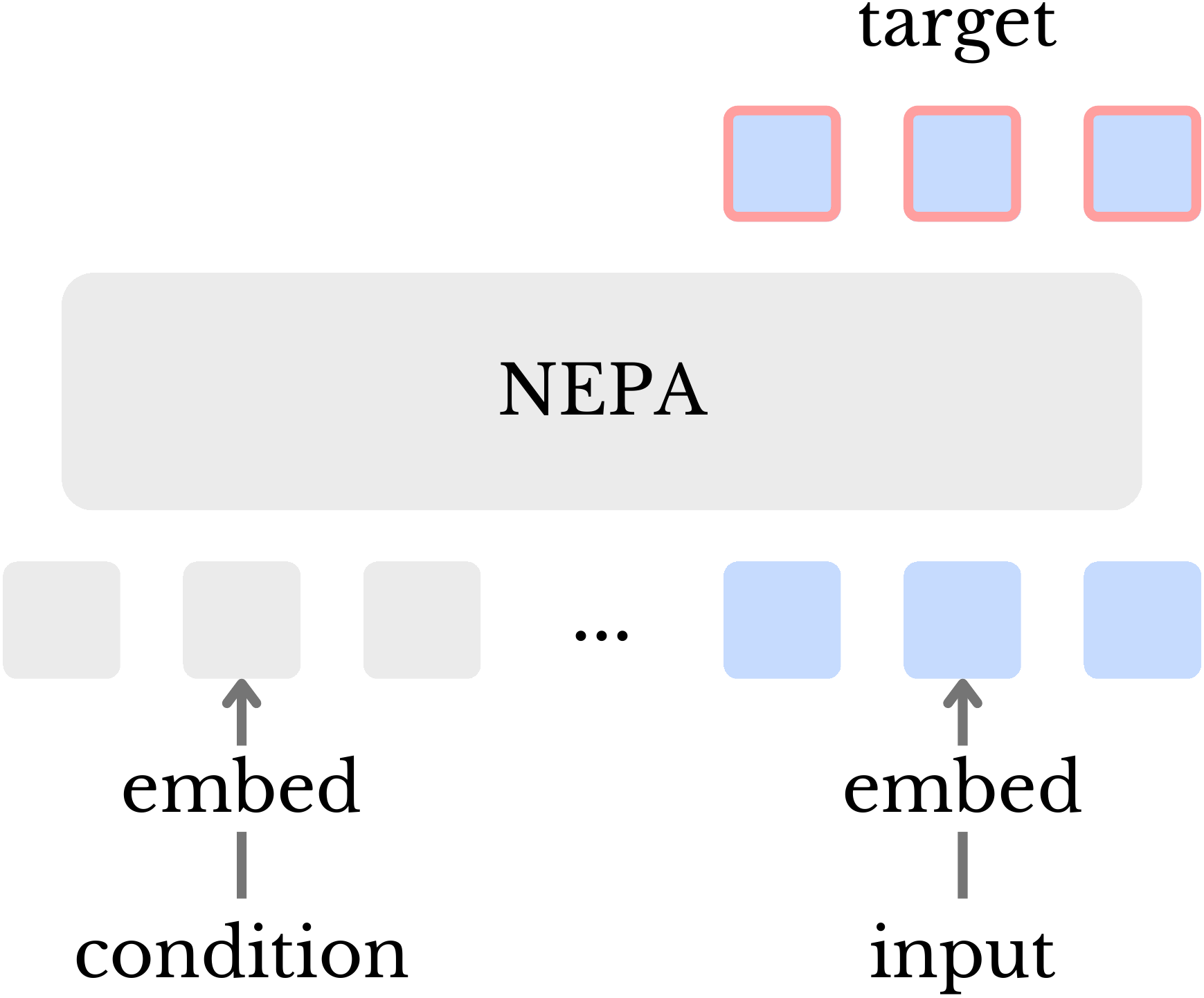}
\vspace{-14pt}
\captionof{figure}{\textbf{MEP for generation.} The clean image embeddings are the next embeddings after the condition and the noisy image; the output at each noisy patch predicts the embedding that follows it.}
\label{fig:nepa}
\end{minipage}\hfill
\begin{minipage}[c]{0.65\textwidth}
Generation produces an image through a sequence of states: from the condition, through a noisy image, to the clean image. We write this as a sequence of embeddings, the condition, then the \(N\) patch embeddings of the noisy image, then the \(N\) patch embeddings of the clean image. In this sequence, the clean embeddings are the next \(N\) embeddings after the condition and the noisy image, and we train a NEPA model to predict them with MEP, taking \(K=N\). The sequence is ordered by generation state rather than by patch position, so the embedding that follows noisy patch \(i\) is clean patch \(i\). MEP predicts all \(N\) embeddings of the clean image in one forward pass, each from the full context of the condition and the noisy image. The model is defined by three choices: the sequence and its readout, an attention mask with the condition as a causal prefix, and an InfoNCE loss.
\end{minipage}
\vspace{4pt}

Given a clean input \(\mathbf{x}_0\), we sample a noise level \(t\) and construct the noisy input \(\mathbf{x}_t = (1-t)\mathbf{x}_0 + t\boldsymbol{\epsilon}\). Both are patchified by the embedding layer into noisy and clean patch embeddings, \(\mathbf{z}_t = f(\mathbf{x}_t)\) and \(\mathbf{z}_0 = f(\mathbf{x}_0)\). As in NEPA, the targets come from the model's own patch embedding layer \(f\), which is learned jointly with the model. The targets are compared through InfoNCE rather than regressed: the model only has to tell each patch apart from the other patches of the image, and the remaining detail is left to the generator. The noise level is sampled uniformly, \(t\sim\mathcal{U}(0,1)\), so that a single model covers every denoising step at inference, and the condition is replaced by a null token with probability 0.1, so that the same model also provides the unconditional predictions used for classifier-free guidance (Section~\ref{sec:ecg}).
As shown in Figure~\ref{fig:nepa}, the NEPA model \(h_\theta\) reads the condition \(c\), a begin-of-image token \(\langle\mathrm{boi}\rangle\), and the noisy embeddings, and predicts the next \(N\) embeddings, the clean image:
\begin{equation}
\label{eq:mep}
\mathbf{p} = h_\theta([c, \langle\mathrm{boi}\rangle], \mathbf{z}_t),
\qquad
\mathcal{L}_{\mathrm{MEP}} = \mathcal{D}(\mathbf{z}_0, \mathbf{p}).
\end{equation}
This is Equation~\ref{eq:mep-general} with the context \([c, \langle\mathrm{boi}\rangle, \mathbf{z}_t]\) and the next \(K=N\) embeddings \(\mathbf{z}_0\). As in NEPA, where the output at each position predicts the embedding that follows it, the output at noisy patch \(i\) predicts its next embedding \(\mathbf{z}_{0,i}\). The noise level sets the difficulty: at \(t=1\) the noisy image carries no information and the clean image must be predicted from the condition alone, whereas as \(t\to0\) the noisy image approaches the clean one.

\vspace{8pt}
\noindent
\begin{minipage}[c]{0.62\textwidth}
\textbf{Attention mask.} The condition tokens and \(\langle\mathrm{boi}\rangle\) form a causal prefix that never attends to image tokens. The image tokens of the noisy image attend to the whole prefix and to each other, and every prediction uses the full spatial context of the noisy image (Figure~\ref{fig:attn}). Since the prefix does not depend on the image, it is encoded once and cached during sampling (Section~\ref{sec:ecg}).

\vspace{5pt}
\textbf{Embedding prediction loss.} Each prediction \(\mathbf{p}_i\) is trained to match its next embedding \(\mathbf{z}_{0,i}\). The distance \(\mathcal{D}\) in Equation~\ref{eq:mep} can take many forms (Section~\ref{sec:ablations}); we use the InfoNCE loss~\citep{oord2018cpc}, with predicted embeddings as queries and clean embeddings as keys.
\end{minipage}\hfill
\begin{minipage}[c]{0.35\textwidth}
\centering
\includegraphics[width=0.95\linewidth]{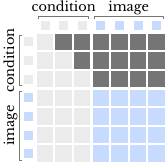}
\vspace{-4pt}
\captionof{figure}{\textbf{Attention mask.} Rows are queries and columns are keys; dark cells are masked.}
\label{fig:attn}
\end{minipage}

\vspace{8pt}
\noindent
\begin{minipage}[c]{0.43\textwidth}
\centering
\captionsetup{justification=centering}
\includegraphics[width=\linewidth]{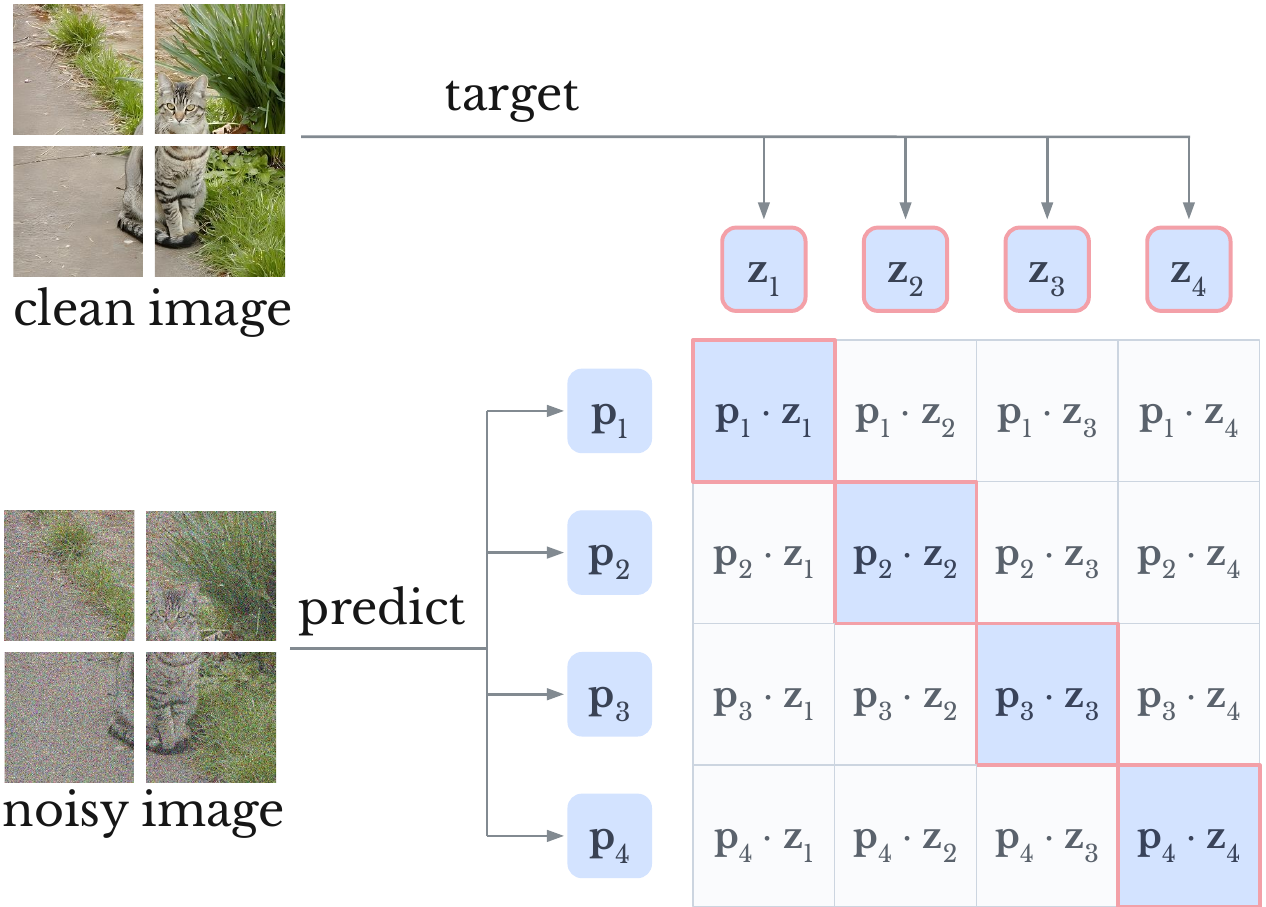}
\vspace{-14pt}
\captionof{figure}{\textbf{Embedding prediction loss.}}
\label{fig:mep-loss}
\end{minipage}\hfill
\begin{minipage}[c]{0.55\textwidth}
As illustrated in Figure~\ref{fig:mep-loss}, the positives lie on the diagonal, pairing each \(\mathbf{p}_i\) with \(\mathbf{z}_{0,i}\), while the other patches of the same image serve as negatives:
\begin{equation}
\label{eq:infonce}
\mathcal{D}(\mathbf{z}_0, \mathbf{p})
=
-\frac{1}{N}\sum_{i=1}^{N} \log
\frac{\exp(\mathbf{p}_i^{\top}\mathbf{z}_{0,i})}
{\sum_{j=1}^{N} \exp(\mathbf{p}_i^{\top}\mathbf{z}_{0,j})}.
\end{equation}
After this stage, the NEPA model maps the condition and the noisy image to predicted clean image embeddings.
\end{minipage}

\Needspace*{4\baselineskip}
\subsection{Embedding Conditioned Generation}\label{sec:ecg}

The second stage trains the generator \(G_\phi\) with the NEPA model \(h_\theta\) frozen as its conditioning module. The NEPA model predicts the clean embeddings, while the generator learns the denoising vector field.

At each denoising step, the current noisy input \(\mathbf{x}_t\) is first fed into the NEPA model, which outputs predictive embeddings \(\mathbf{p}_t = h_\theta([c, \langle\mathrm{boi}\rangle], f(\mathbf{x}_t))\). The generator follows DiT~\citep{peebles2022dit} (details in Appendix~\ref{app:dit-baseline}), except that it is conditioned on \(\mathbf{p}_t\) through the MM-DiT blocks of SD3~\citep{esser2024sd3}: the predicted embeddings and the noisy latent tokens are processed with separate weights and exchange information through joint attention, followed by single-stream blocks on the concatenated sequence. The generator receives the condition only through \(\mathbf{p}_t\) and has no separate class embedding; the timestep enters through adaLN modulation, as in DiT. The training loss is the weighted flow-matching loss

\begin{equation}
\mathcal{L}_{\mathrm{ECG}}
= \mathbb{E}_{\mathbf{x}_0,\boldsymbol{\epsilon},t,c}
\Big[\,w(t)\,\big\|\,\mathbf{v}_t - G_\phi(\mathbf{x}_t, t, \mathbf{p}_t)\,\big\|^2\Big].
\end{equation}

During this stage, gradients update only the generator; the VAE and the NEPA model remain frozen.

\textbf{Condition caching.} Because the condition prefix never attends to image tokens, its key-value cache in the NEPA model is computed once before the denoising loop and reused at every step; only the image tokens are re-encoded as \(\mathbf{x}_t\) changes.

\textbf{Classifier-free guidance}~\citep{ho2022cfg}. During training, we randomly replace the condition \(c\) with a learned null token. At inference, the unconditional branch feeds the null token to the NEPA model, which yields \(\mathbf{p}_t^{\varnothing}\), and guidance is applied to the velocities predicted by the generator:
\begin{equation}
\label{eq:cfg}
\begin{gathered}
\hat{\mathbf{v}}_{\mathrm{cond}} = G_\phi(\mathbf{x}_t, t, \mathbf{p}_t), \qquad
\hat{\mathbf{v}}_{\mathrm{uncond}} = G_\phi(\mathbf{x}_t, t, \mathbf{p}_t^{\varnothing}), \\
\hat{\mathbf{v}}_{\mathrm{cfg}}
=
\hat{\mathbf{v}}_{\mathrm{uncond}}
+
s \cdot
(\hat{\mathbf{v}}_{\mathrm{cond}} - \hat{\mathbf{v}}_{\mathrm{uncond}}),
\end{gathered}
\end{equation}
where \(s\) is the guidance scale.

\Needspace*{4\baselineskip}
\section{Experiments}\label{sec:experiments}

We evaluate our framework on class-conditional ImageNet-1K generation at \(256\times256\) resolution. At small scale, Section~\ref{sec:ablations} ablates the design of Multi-Embedding Prediction and Section~\ref{sec:condition} compares conditions of the generator; Section~\ref{sec:scaling} then scales up the NEPA model and the generator, and Section~\ref{sec:comparison} compares our final model with prior work.

\textbf{Setup.} Following ADM~\citep{dhariwal2021adm}, images are center-cropped to \(256\times256\) and encoded by the frozen SD-VAE~\citep{rombach2021ldm} into \(32\times32\times4\) latents. The NEPA model follows \citet{xu2025nepa} and the generator follows DiT~\citep{peebles2022dit}, each at B, L, and XL scale; NEPA-DiT-B/L/XL denotes a NEPA model paired with a DiT-B/L/XL generator. Unless stated otherwise, the NEPA model is NEPA-XL with patch size 4, and the generator uses patch size 2. The NEPA model is trained with Multi-Embedding Prediction for 240 epochs, then frozen while the generator is trained with the flow-matching loss for 400K iterations (80 epochs), both at batch size 256. Appendix~\ref{app:dit-baseline} gives further details.

\textbf{Evaluation.} We report the Fr\'echet Inception Distance on 50K samples (FID-50K), computed with the ADM evaluation suite and its ImageNet \(256\times256\) reference batch~\citep{dhariwal2021adm}. Unless stated otherwise, ablation and scaling experiments sample with a 64-step ODE solver and no classifier-free guidance; the sampling setup of our final model is given in Section~\ref{sec:comparison}.

\Needspace*{4\baselineskip}
\subsection{Ablation Studies}\label{sec:ablations}

Following previous work~\citep{ma2024sit,wang2025ddt}, we run ablations at the B scale with NEPA-B and patch size 2, one factor at a time (defaults in gray); Table~\ref{tab:abl-patch} then selects patch size 4.

\begin{wraptable}{r}{0.36\textwidth}
\vspace{-12pt}
\centering
\footnotesize
\setlength{\tabcolsep}{5pt}
\begin{tabular}{@{}lcc@{}}
\toprule
Loss & FID\(\downarrow\) & Acc.\ (\%)\(\uparrow\) \\
\midrule
Cos.\ sim. & 37.82 & \textbf{83.1} \\
MSE & 35.23 & 82.6 \\
\rowcolor{defaultrow} InfoNCE & \textbf{31.36} & 81.7 \\
\bottomrule
\end{tabular}
\vspace{-4pt}
\caption{\textbf{Loss.} Acc.\ is ImageNet-1K top-1 accuracy of the fine-tuned NEPA model.}
\label{tab:abl-loss}
\vspace{-10pt}
\end{wraptable}

\textbf{Loss.} We compare the InfoNCE loss of Section~\ref{sec:reasoner} with two regression losses, negative cosine similarity and mean squared error (MSE), which pull each prediction toward its own target without contrasting it against the others. InfoNCE gives the best FID, 31.36 against 35.23 for MSE and 37.82 for cosine similarity (Table~\ref{tab:abl-loss}). ImageNet-1K accuracy of the fine-tuned NEPA model follows the opposite order. InfoNCE must tell each patch apart from the other patches of the same image, so its predictions keep patch-level detail that helps generation; the regression losses favor smoother, more global predictions, which help classification instead.

\noindent
\begin{minipage}[t]{0.46\textwidth}
\vspace{0pt}
\hyphenpenalty=10000\exhyphenpenalty=10000
\begin{minipage}[t]{0.30\linewidth}
\vspace{0pt}
\centering
\footnotesize
\setlength{\tabcolsep}{3pt}
\captionsetup{justification=centering}
\begin{tabular}{@{}lc@{}}
\toprule
Detach & FID\(\downarrow\) \\
\midrule
w/ & n.c. \\
\rowcolor{defaultrow} w/o & \textbf{31.36} \\
\bottomrule
\end{tabular}
\vspace{-4pt}
\captionof{table}{\textbf{Stop-grad.}}
\label{tab:abl-sg}
\end{minipage}\hfill
\begin{minipage}[t]{0.30\linewidth}
\vspace{0pt}
\centering
\footnotesize
\setlength{\tabcolsep}{3pt}
\captionsetup{justification=centering}
\begin{tabular}{@{}lc@{}}
\toprule
Target & FID\(\downarrow\) \\
\midrule
Norm. & 34.76 \\
\rowcolor{defaultrow} Raw & \textbf{31.36} \\
\bottomrule
\end{tabular}
\vspace{-4pt}
\captionof{table}{\textbf{Target norm.}}
\label{tab:abl-norm}
\end{minipage}\hfill
\begin{minipage}[t]{0.35\linewidth}
\vspace{0pt}
\centering
\footnotesize
\setlength{\tabcolsep}{3pt}
\captionsetup{justification=centering}
\begin{tabular}{@{}lc@{}}
\toprule
Timestep & FID\(\downarrow\) \\
\midrule
\rowcolor{defaultrow} w/o & \textbf{31.36} \\
w/ & 31.40 \\
\bottomrule
\end{tabular}
\vspace{-4pt}
\captionof{table}{\textbf{Timestep input.}}
\label{tab:abl-timestep}
\end{minipage}
\end{minipage}\hfill
\begin{minipage}[t]{0.52\textwidth}
\vspace{0pt}
\textbf{Stop-gradient.} Detaching the targets \(\mathbf{z}_0\) from the gradient prevents training from converging (n.c.\ in Table~\ref{tab:abl-sg}), so gradients flow through the targets.

\vspace{5pt}
\textbf{Target normalization.} Normalizing the target embeddings raises FID from 31.36 to 34.76 (Table~\ref{tab:abl-norm}), so we use raw targets.
\end{minipage}

\vspace{5pt}
\textbf{Timestep input.} Feeding the diffusion timestep to the NEPA model changes FID from 31.36 to 31.40 (Table~\ref{tab:abl-timestep}), presumably because the noise level is already evident from the noisy patches; we therefore keep the NEPA model timestep-free, while the generator still receives the timestep through adaLN.

\textbf{Vocabulary.} The InfoNCE loss scores each prediction against a bank of candidate targets. We compare banks built from the patches of the same image (instance), of all images in the per-device batch (batch), and of all images across devices (full). The instance, batch, and full banks give 31.36, 32.13, and 31.21 (Table~\ref{tab:abl-vocab}). We use the instance bank, which requires no communication across devices; its negatives are the other patches of the same image, as in Equation~\ref{eq:infonce}.

\vspace{6pt}
\noindent
\begin{minipage}[t]{0.44\textwidth}
\vspace{0pt}
\footnotesize
\setlength{\tabcolsep}{4pt}
\captionsetup{justification=centering}
\begin{minipage}[t]{0.42\linewidth}
\vspace{0pt}
\centering
\begin{tabular}{@{}lc@{}}
\toprule
Bank & FID\(\downarrow\) \\
\midrule
\rowcolor{defaultrow} Instance & 31.36 \\
Batch & 32.13 \\
Full & \textbf{31.21} \\
\bottomrule
\end{tabular}
\vspace{-4pt}
\captionof{table}{\textbf{Vocabulary.}}
\label{tab:abl-vocab}
\end{minipage}\hfill
\begin{minipage}[t]{0.56\linewidth}
\vspace{0pt}
\centering
\begin{tabular}{@{}ccc@{}}
\toprule
Patch & Tokens & FID\(\downarrow\) \\
\midrule
2 & 256 & 31.36 \\
\rowcolor{defaultrow} 4 & 64 & \textbf{31.00} \\
8 & 16 & 33.20 \\
\bottomrule
\end{tabular}
\vspace{-4pt}
\captionof{table}{\textbf{Patch size.}}
\label{tab:abl-patch}
\end{minipage}
\end{minipage}\hfill
\begin{minipage}[t]{0.53\textwidth}
\vspace{0pt}
\textbf{Patch size.} The patch size of the NEPA model sets the number of predicted embeddings (Table~\ref{tab:abl-patch}). Patch size 4 gives 31.00 against 31.36 for patch size 2 with a quarter of the tokens, which makes the query at every denoising step correspondingly cheaper, whereas patch size 8 is too coarse (33.20). We therefore use patch size 4 in all remaining experiments.
\end{minipage}

\Needspace*{4\baselineskip}
\subsection{Condition of the Generator}\label{sec:condition}

At the same small scale, Table~\ref{tab:app-signal} keeps the generator and its training recipe fixed and varies only what it is conditioned on. The class embedding gives 36.39, and adding the patch embeddings of \(\mathbf{x}_t\) to the class token gives 37.21. The last three rows give the generator the same extra network as ours, of the NEPA-XL architecture, so that their parameters and FLOPs per step match those of our model; what differs is how that network is trained. Trained end to end with the generator as a single network, it gives 29.52. Pretrained as a flow-matching model for 240 epochs, the same budget as MEP, and then frozen, its features give 30.86. Trained with MEP and frozen, its predicted embeddings give 25.04.

\vspace{4pt}
\noindent
\begin{minipage}{\textwidth}
\centering
\footnotesize
\setlength{\tabcolsep}{6pt}
\begin{tabular}{@{}llllc@{}}
\toprule
Condition of the generator & Extra network & Training & Objective & FID\(\downarrow\) \\
\midrule
Class embedding & -- & -- & -- & 36.39 \\
Class token + noisy patches & -- & -- & -- & 37.21 \\
Single network, end to end & NEPA architecture & joint & flow matching & 29.52 \\
Pretrained flow model & NEPA architecture & 240 epochs, frozen & flow matching & 30.86 \\
\rowcolor{defaultrow} NEPA model & NEPA & 240 epochs, frozen & MEP & \textbf{25.04} \\
\bottomrule
\end{tabular}
\vspace{-2pt}
\captionof{table}{\textbf{Condition of the generator.} All rows use the same generator and training recipe (400K iterations, best over sampling steps) and differ only in its condition; the last three rows also carry the extra network of our model.}
\label{tab:app-signal}
\end{minipage}

\vspace{4pt}
\Needspace*{4\baselineskip}
\subsection{Scaling Behavior}\label{sec:scaling}

\noindent
\begin{minipage}[c]{0.460\textwidth}
\centering
\footnotesize
\setlength{\tabcolsep}{4pt}
\begin{tabular}{@{}lcccc@{}}
\toprule
& & \multicolumn{3}{c}{Generator (FID\(\downarrow\))} \\
\cmidrule(l){3-5}
NEPA & Acc.\(\uparrow\) & DiT-B & DiT-L & DiT-XL \\
\midrule
NEPA-B & 81.3 & 31.00 & 17.69 & 16.54 \\
NEPA-L & 83.4 & 27.46 & 14.35 & 13.74 \\
NEPA-XL & \textbf{85.2} & 25.04 & 12.86 & \textbf{11.27} \\
\bottomrule
\end{tabular}
\vspace{-2pt}
\captionof{table}{\textbf{Scaling.} FID-50K after 400K iterations; Acc.\ is ImageNet-1K top-1 accuracy (\%) of the fine-tuned NEPA model.}
\label{tab:scaling}
\end{minipage}\hfill
\begin{minipage}[c]{0.500\textwidth}
\centering
\includegraphics[width=0.95\linewidth]{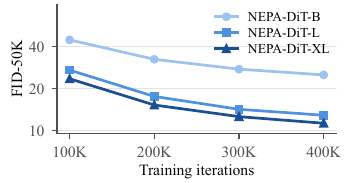}
\vspace{-8pt}
\captionof{figure}{\textbf{Training progress} of NEPA-DiT-B/L/XL, FID-50K without guidance.}
\label{fig:training-scaling}
\end{minipage}

\vspace{6pt}
\textbf{Scaling the NEPA model and the generator.} Beyond the small-scale setting, Table~\ref{tab:scaling} scales both models, training all nine combinations of NEPA-B/L/XL and DiT-B/L/XL, each for 400K iterations and evaluated with 64 ODE steps without guidance. FID decreases monotonically along both axes. Scaling the generator from DiT-B to DiT-XL improves FID by about 14 points for every NEPA model, and scaling the NEPA model from NEPA-B to NEPA-XL improves FID by about 5 points for every generator. All entries use patch size 4 for the NEPA model and patch size 2 for the generator, and the generators match DiT-B/L/XL in depth, width, and number of heads. Each NEPA model is trained once and shared by the three generators of its row.

\textbf{Training progress.} Figure~\ref{fig:training-scaling} follows NEPA-DiT-B/L/XL over training; all three use the same frozen NEPA-XL model. Each point is FID-50K with 64 ODE steps and no guidance, evaluated every 100K iterations. Larger generators are better at every checkpoint, and the gap opens early: at 200K iterations, NEPA-DiT-L (17.55) and NEPA-DiT-XL (15.23) are already well ahead of NEPA-DiT-B at 400K (25.04). At 400K iterations, the three reach 25.04, 12.86, and 11.27, the NEPA-XL row of Table~\ref{tab:scaling}.

\Needspace*{4\baselineskip}
\subsection{Comparison with Previous Methods}\label{sec:comparison}

\begin{table}[t]
\centering
\caption{\textbf{Class-conditional generation on ImageNet \(256\times256\)} with guidance. Top (gray): methods with modified or alternative tokenizers. Bottom: methods in the standard SD-VAE latent space. Metrics and training epochs on ImageNet-1K are taken from each paper; for ours, epochs are those of the NEPA model + the generator. ``--'' means not reported. \(^\ddagger\)AutoGuidance.}
\label{tab:system}
\footnotesize
\setlength{\tabcolsep}{2.8pt}
\resizebox{\linewidth}{!}{%
\begin{tabular}{@{}llcccccc@{}}
\toprule
Method & Tokenizer & Epochs & FID\(\downarrow\) & sFID\(\downarrow\) & IS\(\uparrow\) & Pre.\(\uparrow\) & Rec.\(\uparrow\) \\
\midrule
\multicolumn{8}{@{}l}{\gr{\textbf{Other tokenizers}}} \\
\midrule
\gr{REPA + EQ-VAE~\citep{kouzelis2025eqvae}} & \gr{EQ-VAE} & \gr{200} & \gr{1.70} & \gr{5.13} & \gr{283.0} & \gr{0.79} & \gr{0.62} \\
\gr{LightningDiT-XL/1~\citep{yao2025lightningdit}} & \gr{VA-VAE} & \gr{800} & \gr{1.35} & \gr{4.15} & \gr{295.3} & \gr{0.79} & \gr{0.65} \\
\gr{LightningDiT + IG~\citep{zhou2026internalguidance}} & \gr{VA-VAE} & \gr{680} & \gr{1.19} & \gr{4.11} & \gr{269.0} & \gr{0.79} & \gr{0.66} \\
\gr{DiT-XL + CMuon~\citep{chen2026cmuon}} & \gr{VA-VAE} & \gr{200} & \gr{1.18} & \gr{--} & \gr{--} & \gr{--} & \gr{--} \\
\gr{REPA-E~\citep{leng2025repae}} & \gr{E2E-VAE} & \gr{800} & \gr{1.12} & \gr{4.09} & \gr{302.9} & \gr{0.79} & \gr{0.66} \\
\gr{Send-VAE + REPA~\citep{page2026sendvae}} & \gr{Send-VAE} & \gr{800} & \gr{1.21} & \gr{4.10} & \gr{315.1} & \gr{0.79} & \gr{0.66} \\
\gr{SFD-XL\(^\ddagger\)~\citep{pan2026sfd}} & \gr{SemVAE} & \gr{800} & \gr{1.06} & \gr{3.89} & \gr{267.0} & \gr{0.78} & \gr{0.67} \\
\gr{RAE DiT\(^\text{DH}\)-XL\(^\ddagger\)~\citep{zheng2025rae}} & \gr{RAE} & \gr{800} & \gr{1.13} & \gr{--} & \gr{262.6} & \gr{0.78} & \gr{0.67} \\
\gr{MixFlow + RAE~\citep{li2026mixflow}} & \gr{RAE} & \gr{800+200} & \gr{1.10} & \gr{4.40} & \gr{259.7} & \gr{0.78} & \gr{0.67} \\
\gr{RAEv2~\citep{singh2026raev2}} & \gr{RAE} & \gr{80} & \gr{1.06} & \gr{--} & \gr{255.3} & \gr{--} & \gr{--} \\
\midrule
\multicolumn{8}{@{}l}{\textbf{SD-VAE}} \\
\midrule
DiT-XL/2~\citep{peebles2022dit} & SD-VAE & 1400 & 2.27 & 4.60 & 278.2 & 0.83 & 0.57 \\
SiT-XL/2~\citep{ma2024sit} & SD-VAE & 1400 & 2.06 & 4.49 & 277.5 & 0.83 & 0.59 \\
REPA~\citep{yu2024repa} & SD-VAE & 800 & 1.42 & 4.70 & 305.7 & 0.80 & 0.65 \\
TREAD~\citep{krause2025tread} & SD-VAE & 740 & 1.69 & 4.73 & 292.7 & 0.81 & 0.63 \\
DDT-XL/2~\citep{wang2025ddt} & SD-VAE & 400 & 1.26 & -- & 310.6 & 0.79 & 0.65 \\
REG~\citep{wu2025reg} & SD-VAE & 800 & 1.36 & 4.25 & 299.4 & 0.77 & 0.66 \\
U-REPA~\citep{tian2025urepa} & SD-VAE & 400 & 1.41 & -- & -- & -- & -- \\
SRA~\citep{jiang2025sra} & SD-VAE & 800 & 1.58 & 4.65 & 311.4 & 0.80 & 0.63 \\
LSEP~\citep{yun2025lsep} & SD-VAE & 800 & 1.46 & 4.94 & 296.8 & 0.80 & 0.64 \\
SPRINT + REPA~\citep{park2025sprint} & SD-VAE & 400 & 1.59 & -- & -- & 0.80 & 0.64 \\
SiT-XL/2 + IG~\citep{zhou2026internalguidance} & SD-VAE & 800 & 1.46 & 4.79 & 265.7 & 0.80 & 0.64 \\
SiT-XL/2 + Sparse Guid.~\citep{krause2026sparseguidance} & SD-VAE & 400 & 1.58 & 4.45 & 249.7 & 0.80 & 0.63 \\
SRA 2~\citep{wang2026sra2} & SD-VAE & 800 & 1.52 & 4.63 & 316.2 & 0.82 & 0.62 \\
UDT\(^+\)-XL/2 + REPA~\citep{yun2026udt} & SD-VAE & 320 & 1.38 & 4.38 & 306.3 & 0.79 & 0.66 \\
RecFM-XL~\citep{huang2026recfm} & SD-VAE & 160 & 2.49 & -- & -- & -- & -- \\
\rowcolor{oursrow} \textbf{NEPA-DiT-XL + REPA (ours)}, ODE-96 & SD-VAE & 240+80 & 1.57 & 4.82 & 298.0 & 0.79 & 0.63 \\
\rowcolor{oursrow} \textbf{NEPA-DiT-XL + REPA (ours)}, SDE-250 & SD-VAE & 240+80 & 1.32 & 4.32 & 311.0 & 0.79 & 0.68 \\
\bottomrule
\end{tabular}}
\end{table}

\begin{wraptable}{r}{0.33\textwidth}
\vspace{-12pt}
\centering
\footnotesize
\setlength{\tabcolsep}{3pt}
\begin{tabular}{@{}cccc@{}}
\toprule
Scale & Interval & ODE-96 & SDE-250 \\
\midrule
1.0 & \([0,1]\) & 8.77 & 7.59 \\
1.4 & \([0,1]\) & 2.58 & 2.32 \\
2.4 & \([0.3,1]\) & 1.72 & 1.46 \\
\rowcolor{defaultrow} 3.6 & \([0.4,1]\) & \textbf{1.57} & \textbf{1.32} \\
\bottomrule
\end{tabular}
\vspace{-4pt}
\caption{\textbf{Guidance.} Applied only for \(t\) in the interval. Shaded: setting of Table~\ref{tab:system}.}
\label{tab:abl-cfg}
\vspace{-10pt}
\end{wraptable}
\textbf{Setup.} Our final model combines Embedding Conditioned Generation, which sets the condition of the generator, with REPA~\citep{yu2024repa}, which aligns its features. NEPA-DiT-XL pairs NEPA-XL, the frozen model of Section~\ref{sec:scaling}, with a DiT-XL generator, and the generator is trained with both the flow-matching loss and the REPA alignment loss (DINOv2-B aligned at the third two-stream block, \(\lambda=0.5\)) for 400K iterations. We report results with the 96-step ODE and the 250-step SDE samplers of SiT~\citep{ma2024sit}. Besides FID~\citep{heusel2017fid}, we report sFID~\citep{nash2021sfid}, Inception Score~\citep{salimans2016improved}, and precision and recall~\citep{kynkaanniemi2019precision} on the same 50K samples.

\begin{figure}[t]
\centering
\begin{minipage}{\linewidth}
\centering
\includegraphics[width=\linewidth]{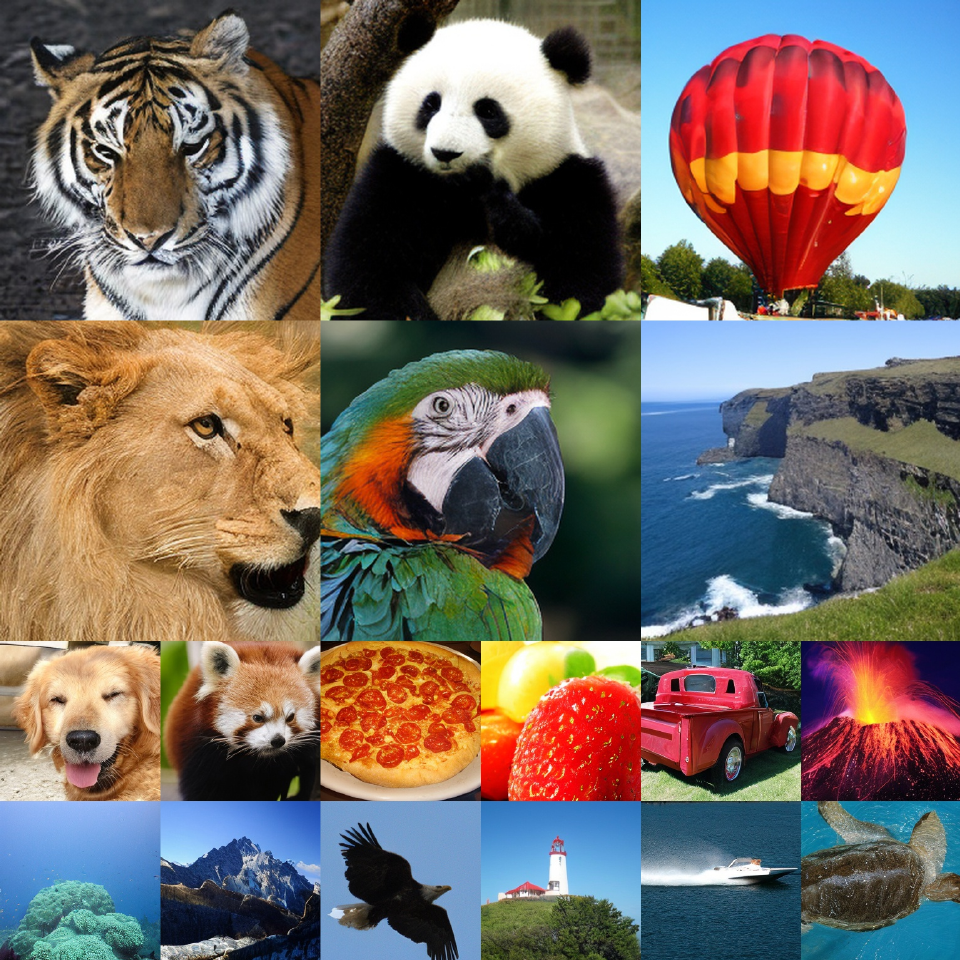}
\caption{\textbf{Qualitative results} of NEPA-DiT-XL on ImageNet \(256\times256\).}
\label{fig:samples}
\end{minipage}
\end{figure}

\textbf{Guidance.} Table~\ref{tab:abl-cfg} lists guidance settings. Restricting guidance to an interval~\citep{kynkaanniemi2024interval} admits larger scales: scale 3.6 on \(t\in[0.4,1]\) gives 1.57 and 1.32, the best with both samplers, selected by FID-50K. All guided results in Table~\ref{tab:system} use this setting.

\textbf{Query rate.} With the 96-step ODE sampler, computing the NEPA predictions only once, at the first step, and reusing them for all later steps gives an FID of 242.63, as all later steps are then conditioned on embeddings predicted from pure noise. Querying every 8 steps gives 1.82, against 1.57 at every step (Table~\ref{tab:app-refresh}). Querying at every step also sets the sampling cost of the NEPA model, 92 GFLOPs per step next to 309 for the generator; with 96 ODE steps, one image takes 62 TFLOPs in all, below the 91 TFLOPs of SiT-XL/2 with REPA with 250 SDE steps.

\textbf{Results.} Among the methods of Table~\ref{tab:system} in the SD-VAE latent space, NEPA-DiT-XL + REPA reaches an FID of 1.32 with the 250-step SDE sampler and 1.57 with the 96-step ODE sampler. Most of its training compute goes into the NEPA model, trained once: with it included, NEPA-DiT-XL, 1.39B parameters in all, is trained with \(3.1\times10^{20}\) FLOPs, about a third of what SiT-XL/2 with REPA uses over its 800 epochs to reach 1.42 (Appendix~\ref{app:dit-baseline}). Figure~\ref{fig:samples} and Appendix~\ref{app:samples} show samples.

\Needspace*{4\baselineskip}
\section{Conclusion}
We have shown that predictive embeddings can serve as the condition of an image generator. In Embedding Conditioned Generation, a DiT generator is conditioned at every denoising step on the clean-image embeddings that a NEPA model, trained with Multi-Embedding Prediction, predicts from the condition and the current noisy image. On ImageNet \(256\times256\), combined with REPA, NEPA-DiT-XL reaches an FID of 1.32 with the 250-step SDE sampler and 1.57 with the 96-step ODE sampler, trained with about a third of REPA's compute even though it trains and runs two networks. Code and model weights will be released. Future work is discussed in Appendix~\ref{app:discussion}.

\Needspace*{4\baselineskip}
\subsection*{AI use statement}

We used large language models only to polish the writing of this paper, such as correcting grammar and improving wording. The authors reviewed all edits and take full responsibility for the content.

{
    \small
    \catcode`\&=12
    \def\textbackslash alpha{\ensuremath{\alpha}}
    \bibliographystyle{style_iclr2027/iclr2027_compact}
    \bibliography{main}
}

\clearpage
\appendix
\FloatBarrier
\Needspace*{4\baselineskip}
\section{Relation to Other Methods}\label{app:conditioning-designs}

\textbf{Relation to RepFusion.} RepFusion~\citep{pan2026repfusion} also conditions a diffusion transformer on the outputs of a model that reads the noisy input at every denoising step. It builds on two existing components: the latent space of a representation autoencoder~\citep{zheng2025rae}, which it adopts so that noisy latents are compatible with pretrained language-model priors, and a pretrained multimodal LLM, which it repurposes as the encoder of noisy representations. Our NEPA model depends on neither. It is trained from scratch with Multi-Embedding Prediction directly on the latents of a standard VAE, so the conditioning interface is defined by a training objective rather than by an existing model. The same recipe can therefore be applied to any Transformer as the NEPA model and to any generator that accepts conditioning tokens.

\textbf{Relation to DDT.} DDT~\citep{wang2025ddt} also splits the generator into a condition encoder that reads the noisy latent and a velocity decoder conditioned on the encoder output. The two are trained jointly: the encoder is supervised indirectly through the velocity loss of the decoder, together with a REPA alignment to DINOv2 features. Our NEPA model is instead trained separately to predict the clean embeddings of the image and is then frozen, so the condition is a representation learned with its own objective rather than an intermediate feature of the generator. DDT also shares the encoder output between adjacent denoising steps to accelerate sampling, and Table~\ref{tab:app-refresh} studies the corresponding choice for our NEPA model.

\textbf{Relation to \(\mathbf{x}_0\)-prediction and self-conditioning.} A diffusion model can be parameterized to predict the clean sample \(\mathbf{x}_0\)~\citep{ho2020ddpm,karras2022edm}, and JiT~\citep{li2026jit} predicts it directly with a Transformer on raw pixels. Self-conditioning~\citep{chen2023analogbits} and RIN~\citep{jabri2023rin} feed such an estimate, or the latent state behind it, back into the same denoiser at the next step. In both cases the estimate comes from the denoiser itself and is trained by its denoising loss. Our NEPA model is a separate network, trained with its own embedding prediction objective and then frozen, and its predictions are recomputed from the current noisy input at every step. Table~\ref{tab:app-signal} includes, in the same setting, a network of the same architecture trained end to end with the generator by flow matching, and the same network pretrained as a flow-matching model and then frozen.

\Needspace*{4\baselineskip}
\section{Future Work}\label{app:discussion}

\textbf{Text conditions.} Text-to-image models either encode the prompt once with a separate text encoder~\citep{saharia2022imagen,chen2023pixart,esser2024sd3}, which cannot adapt to the evolving noisy image and often fixes the length of the condition, or process the prompt together with the image in a single unified model~\citep{deng2025emerging}, where the condition and the image share one backbone. A NEPA model could instead read the whole prompt before the noisy image and predict the clean image embeddings from both, as it does for class labels in this paper.

\textbf{Unified models.} Since the NEPA model is an autoregressive Transformer, a single NEPA model could in principle serve both as a language model with an output head and as the conditioning model of a DiT generator. Studying text conditions, other latent spaces, and higher resolutions are directions for future work.

\Needspace*{4\baselineskip}
\section{Hyperparameter and More Implementation Details}\label{app:dit-baseline}

\begin{table}[h]
\centering
\small
\setlength{\tabcolsep}{8pt}
\renewcommand{\arraystretch}{1.1}
\caption{\textbf{Hyperparameter setup of the NEPA model.}}
\label{tab:hparams-reasoner}
\begin{tabular}{@{}lccc@{}}
\toprule
 & NEPA-B & NEPA-L & NEPA-XL \\
\midrule
\textit{Architecture} & & & \\
Input dim. & \(32\times32\times4\) & \(32\times32\times4\) & \(32\times32\times4\) \\
Patch size & 4 & 4 & 4 \\
Num.\ layers & 12 & 24 & 28 \\
Hidden dim. & 768 & 1,024 & 1,152 \\
Num.\ heads & 12 & 16 & 16 \\
\midrule
\textit{Interpolants} & & & \\
\(\alpha_t\) & \(1-t\) & \(1-t\) & \(1-t\) \\
\(\sigma_t\) & \(t\) & \(t\) & \(t\) \\
\(t\) & \(\mathcal{U}(0,1)\) & \(\mathcal{U}(0,1)\) & \(\mathcal{U}(0,1)\) \\
Training objective & InfoNCE & InfoNCE & InfoNCE \\
\midrule
\textit{Optimization} & & & \\
Training epochs & 240 & 240 & 240 \\
Batch size & 256 & 256 & 256 \\
Optimizer & AdamW & AdamW & AdamW \\
lr & 0.0001 & 0.0001 & 0.0001 \\
\((\beta_1,\beta_2)\) & (0.9, 0.95) & (0.9, 0.95) & (0.9, 0.95) \\
Weight decay & 0.05 & 0.05 & 0.05 \\
lr schedule & cosine & cosine & cosine \\
Warm-up steps & 2K & 2K & 2K \\
Condition dropout & 0.1 & 0.1 & 0.1 \\
\bottomrule
\end{tabular}
\end{table}

\begin{table}[h]
\centering
\small
\setlength{\tabcolsep}{6pt}
\renewcommand{\arraystretch}{1.1}
\caption{\textbf{Hyperparameter setup of the generator.} Num.\ layers counts two-stream + single-stream MM-DiT blocks. The last column is the model in Section~\ref{sec:comparison}.}
\label{tab:hparams-generator}
\begin{tabular}{@{}lcccc@{}}
\toprule
 & NEPA-DiT-B & NEPA-DiT-L & NEPA-DiT-XL & NEPA-DiT-XL + REPA \\
\midrule
\textit{Architecture} & & & & \\
Input dim. & \(32\times32\times4\) & \(32\times32\times4\) & \(32\times32\times4\) & \(32\times32\times4\) \\
Patch size & 2 & 2 & 2 & 2 \\
Num.\ layers & 4+8 & 8+16 & 10+18 & 10+18 \\
Hidden dim. & 768 & 1,024 & 1,152 & 1,152 \\
Num.\ heads & 12 & 16 & 16 & 16 \\
NEPA model & NEPA-XL & NEPA-XL & NEPA-XL & NEPA-XL \\
\midrule
\textit{REPA} & & & & \\
\(\lambda\) & - & - & - & 0.5 \\
Alignment depth & - & - & - & 3 (two-stream) \\
\(\mathrm{sim}(\cdot,\cdot)\) & - & - & - & cos.\ sim. \\
Encoder & - & - & - & DINOv2-B (reg.) \\
\midrule
\textit{Optimization} & & & & \\
Training iteration & 400K & 400K & 400K & 400K \\
Batch size & 256 & 256 & 256 & 256 \\
Optimizer & AdamW & AdamW & AdamW & AdamW \\
lr & 0.0001 & 0.0001 & 0.0001 & 0.0001 \\
\((\beta_1,\beta_2)\) & (0.9, 0.999) & (0.9, 0.999) & (0.9, 0.999) & (0.9, 0.999) \\
Weight decay & 0 & 0 & 0 & 0 \\
Condition dropout & 0.1 & 0.1 & 0.1 & 0.1 \\
\midrule
\textit{Interpolants} & & & & \\
\(\alpha_t\) & \(1-t\) & \(1-t\) & \(1-t\) & \(1-t\) \\
\(\sigma_t\) & \(t\) & \(t\) & \(t\) & \(t\) \\
\(t\) & logit-normal & logit-normal & logit-normal & logit-normal \\
Training objective & v-prediction & v-prediction & v-prediction & v-prediction \\
\bottomrule
\end{tabular}
\end{table}

\textbf{Further implementation details.} Throughout the experiments, the generator matches DiT~\citep{peebles2022dit} in depth, width, and number of heads, but uses MM-DiT blocks~\citep{esser2024sd3}: the first blocks are two-stream and process the predicted embeddings and the latent tokens with separate weights, and the remaining blocks are single-stream blocks on the concatenated sequence. The NEPA model is kept frozen when training the generator. We use mixed precision (bfloat16) with gradient clipping at 1.0 for both stages, and keep an EMA of the generator weights with decay 0.9999. The timestep of the generator is sampled from a logit-normal distribution with zero mean and unit variance~\citep{esser2024sd3}. We use the \texttt{stabilityai/sd-vae-ft-mse} VAE~\citep{rombach2021ldm} for encoding images during training and the \texttt{stabilityai/sd-vae-ft-ema} VAE for decoding samples. For the REPA projector, we use a three-layer MLP with SiLU activations and hidden dimension 2,048, following \citet{yu2024repa}. We provide a detailed hyperparameter setup in Tables~\ref{tab:hparams-reasoner} and~\ref{tab:hparams-generator}.

\textbf{Pretrained encoder.} For REPA, we use the official DINOv2-B checkpoint with registers.

\textbf{ImageNet-1K accuracy.} The accuracy reported in Tables~\ref{tab:abl-loss} and~\ref{tab:scaling} is the top-1 accuracy of the NEPA model after fine-tuning it for ImageNet-1K classification with the fine-tuning recipe of NEPA~\citep{xu2025nepa}.

\textbf{Sampling.} For the system-level comparison, we use the SDE Euler-Maruyama sampler of SiT~\citep{ma2024sit} with 250 steps, or the Euler ODE sampler with 96 steps, and apply classifier-free guidance with a scale of 3.6 only for \(t\in[0.4,1]\)~\citep{kynkaanniemi2024interval}. Other results use the Euler ODE sampler, and the number of steps is given with each table.

\textbf{DiT baseline.} The DiT-B baseline uses the same generator as NEPA-DiT-B, but conditions it directly on a learned class embedding instead of the embeddings predicted by the NEPA model. The class embedding enters the text stream of the MM-DiT blocks as a single token, in place of the predicted embeddings; in both models, the timestep enters through adaLN modulation. All other settings, including the training objective, timestep distribution, and optimization, are the same (Table~\ref{tab:hparams-generator}).

\textbf{Computing resources.} All models are trained on NVIDIA H200 GPUs. The NEPA model is trained once for 240 epochs and is then frozen and shared by all generators; each generator is trained for 400K iterations at batch size 256 (80 epochs). Table~\ref{tab:cost} lists the parameters and the sampling FLOPs of both models, counted with the PyTorch FLOP counter for one \(256\times256\) image (one multiply-add counts as two FLOPs); SiT-XL/2 is counted in the same way. Training FLOPs are counted with the same counter and include on-the-fly VAE encoding: the NEPA model (240 epochs) and the generator with REPA (400K iterations) together take \(3.1\times10^{20}\) FLOPs, and SiT-XL/2 with REPA takes about \(1.0\times10^{21}\) FLOPs for its 800 epochs. With classifier-free guidance, the generator and the NEPA model both run on the conditional and the unconditional branch at each guided step.

\begin{table}[h]
\centering
\small
\setlength{\tabcolsep}{6pt}
\caption{\textbf{Parameters and sampling FLOPs.} GFLOPs per sampling step for one image; the NEPA model runs on the 64 image tokens with the condition cached. Totals include guidance on \(t\in[0.4,1]\) and VAE decoding. SiT-XL/2 is counted in the same way.}
\label{tab:cost}
\begin{tabular}{@{}lccccc@{}}
\toprule
& \multicolumn{2}{c}{Params} & \multicolumn{2}{c}{GFLOPs / step} & \\
\cmidrule(lr){2-3}\cmidrule(lr){4-5}
Model & Generator & NEPA & Generator & NEPA & TFLOPs / image \\
\midrule
NEPA-DiT-B & 135M & 711M & 59 & 92 & -- \\
NEPA-DiT-L & 458M & 711M & 210 & 92 & -- \\
NEPA-DiT-XL, ODE-96 & 683M & 711M & 309 & 92 & 62 \\
NEPA-DiT-XL, SDE-250 & 683M & 711M & 309 & 92 & 160 \\
SiT-XL/2 + REPA, SDE-250 & 675M & -- & 229 & -- & 91 \\
\bottomrule
\end{tabular}
\end{table}

\Needspace*{4\baselineskip}
\section{Additional Ablations}\label{app:more-ablations}

\noindent
\begin{minipage}[t]{0.40\textwidth}
\vspace{0pt}
\centering
\small
\setlength{\tabcolsep}{6pt}
\begin{tabular}{@{}lc@{}}
\toprule
NEPA query & FID\(\downarrow\) \\
\midrule
\rowcolor{defaultrow} Every step & \textbf{1.57} \\
Every 8 steps & 1.82 \\
Once, at the first step & 242.63 \\
\bottomrule
\end{tabular}
\vspace{-2pt}
\captionof{table}{\textbf{Refresh rate.} NEPA-DiT-XL + REPA, 96-step ODE sampling with the guidance of Table~\ref{tab:system}. The generator runs at every step; between queries, it reuses the last predicted embeddings.}
\label{tab:app-refresh}
\end{minipage}\hfill
\begin{minipage}[t]{0.56\textwidth}
\vspace{0pt}
\textbf{Refresh rate.} Table~\ref{tab:app-refresh} queries the NEPA model less often than every denoising step, reusing the last prediction in between. Querying it every 8 steps raises FID from 1.57 to 1.82, and querying it only once, at the first step, gives 242.63.
\end{minipage}

\clearpage
\Needspace*{4\baselineskip}
\section{Additional Qualitative Results}\label{app:samples}
\begin{figure}[!h]
\centering
\begin{subfigure}{0.47\linewidth}
\includegraphics[width=\linewidth]{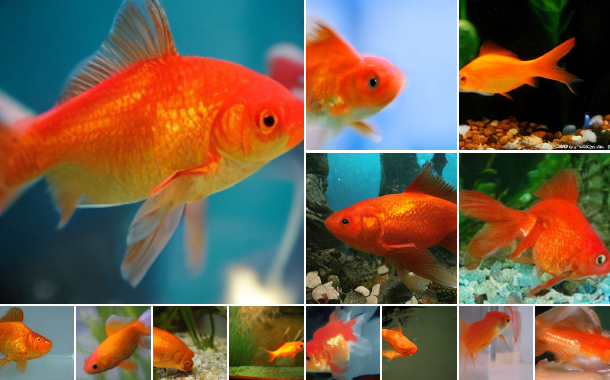}
\caption{Goldfish (1)}
\end{subfigure}\hspace{6pt}
\begin{subfigure}{0.47\linewidth}
\includegraphics[width=\linewidth]{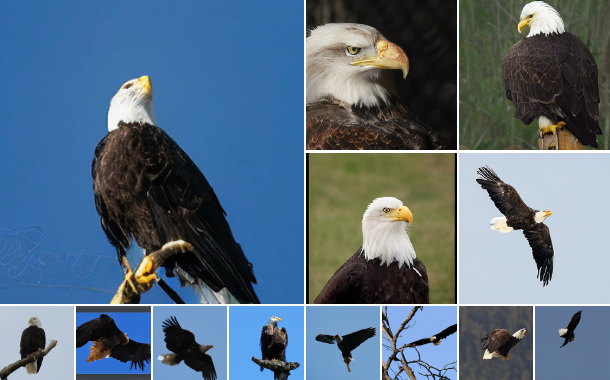}
\caption{Bald eagle (22)}
\end{subfigure}\\[4pt]
\begin{subfigure}{0.47\linewidth}
\includegraphics[width=\linewidth]{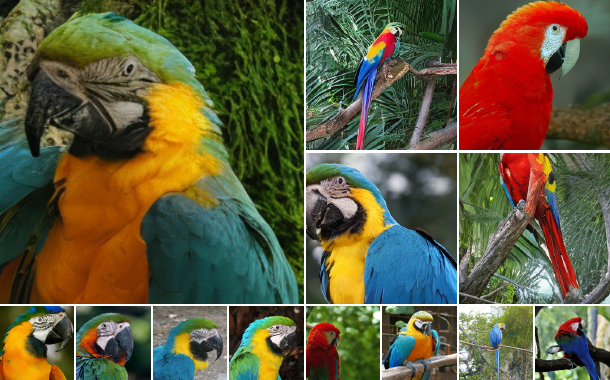}
\caption{Macaw (88)}
\end{subfigure}\hspace{6pt}
\begin{subfigure}{0.47\linewidth}
\includegraphics[width=\linewidth]{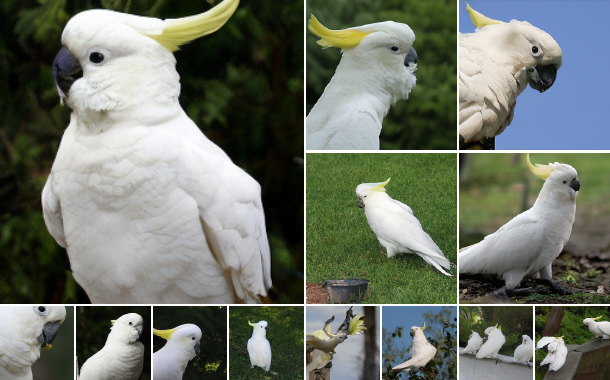}
\caption{Sulphur-crested cockatoo (89)}
\end{subfigure}\\[4pt]
\begin{subfigure}{0.47\linewidth}
\includegraphics[width=\linewidth]{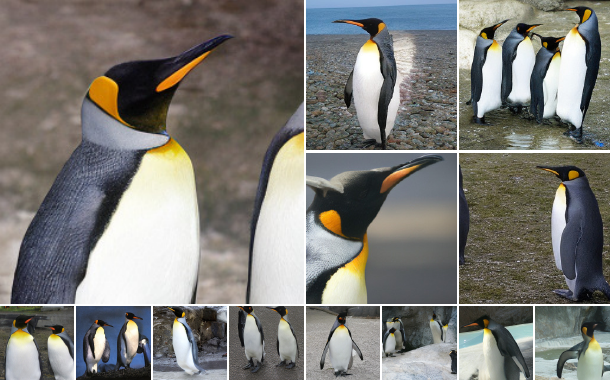}
\caption{King penguin (145)}
\end{subfigure}\hspace{6pt}
\begin{subfigure}{0.47\linewidth}
\includegraphics[width=\linewidth]{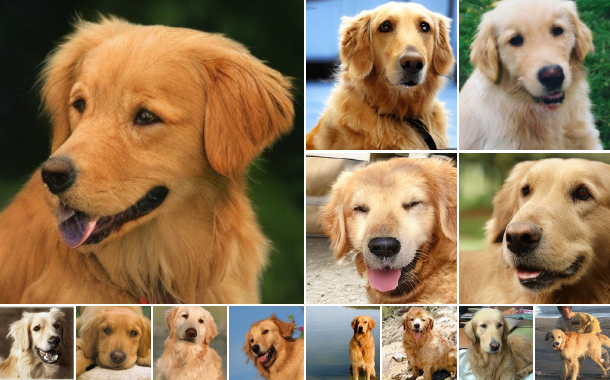}
\caption{Golden retriever (207)}
\end{subfigure}\\[4pt]
\begin{subfigure}{0.47\linewidth}
\includegraphics[width=\linewidth]{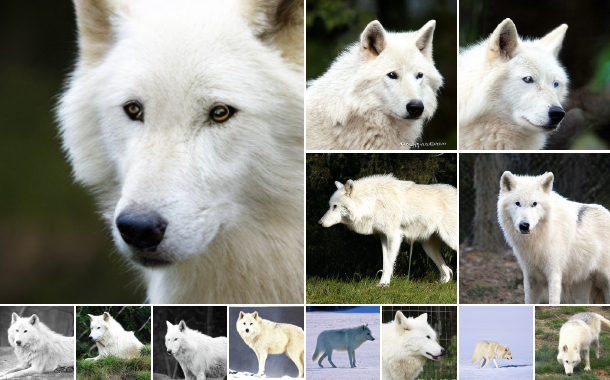}
\caption{White wolf (270)}
\end{subfigure}\hspace{6pt}
\begin{subfigure}{0.47\linewidth}
\includegraphics[width=\linewidth]{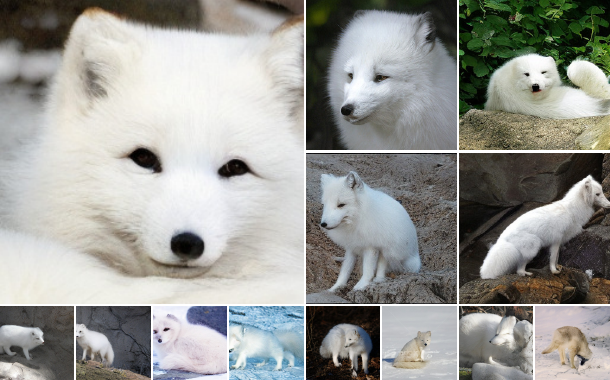}
\caption{Arctic fox (279)}
\end{subfigure}
\caption{\textbf{Samples of NEPA-DiT-XL} on ImageNet \(256\times256\).}
\label{fig:samples-1}
\end{figure}

\begin{figure}[t]
\centering
\begin{subfigure}{0.47\linewidth}
\includegraphics[width=\linewidth]{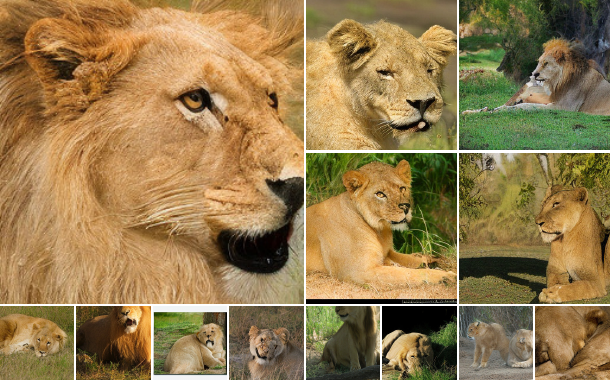}
\caption{Lion (291)}
\end{subfigure}\hspace{6pt}
\begin{subfigure}{0.47\linewidth}
\includegraphics[width=\linewidth]{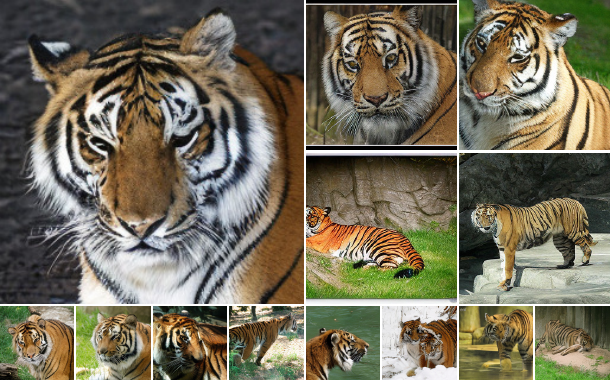}
\caption{Tiger (292)}
\end{subfigure}\\[4pt]
\begin{subfigure}{0.47\linewidth}
\includegraphics[width=\linewidth]{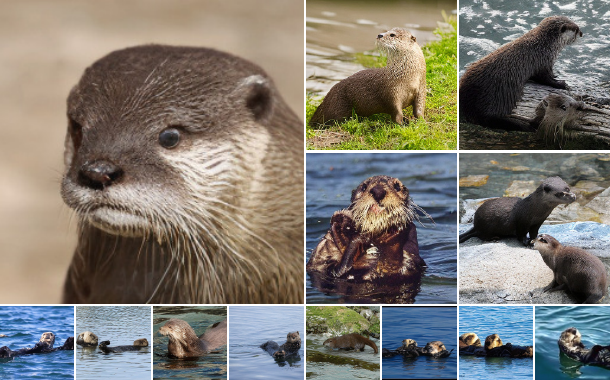}
\caption{Otter (360)}
\end{subfigure}\hspace{6pt}
\begin{subfigure}{0.47\linewidth}
\includegraphics[width=\linewidth]{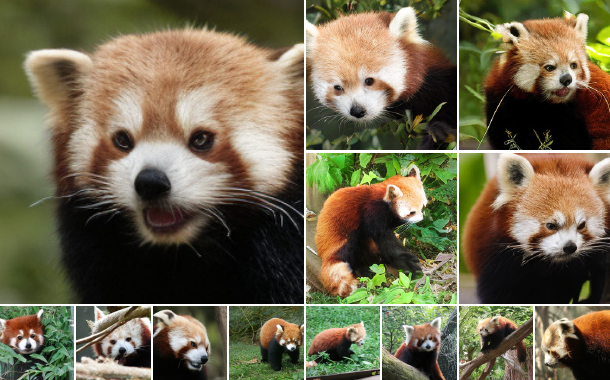}
\caption{Lesser panda (387)}
\end{subfigure}\\[4pt]
\begin{subfigure}{0.47\linewidth}
\includegraphics[width=\linewidth]{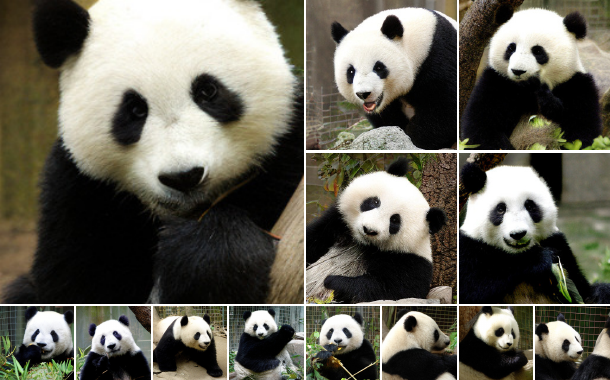}
\caption{Giant panda (388)}
\end{subfigure}\hspace{6pt}
\begin{subfigure}{0.47\linewidth}
\includegraphics[width=\linewidth]{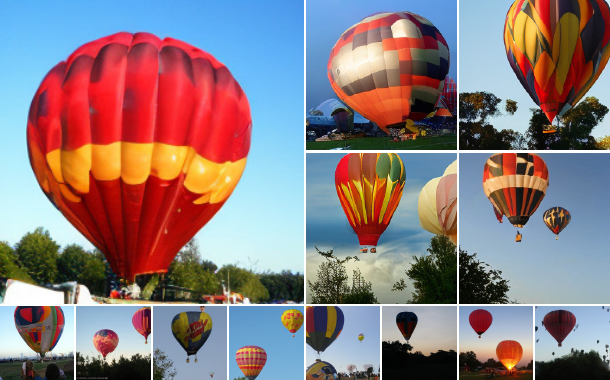}
\caption{Balloon (417)}
\end{subfigure}\\[4pt]
\begin{subfigure}{0.47\linewidth}
\includegraphics[width=\linewidth]{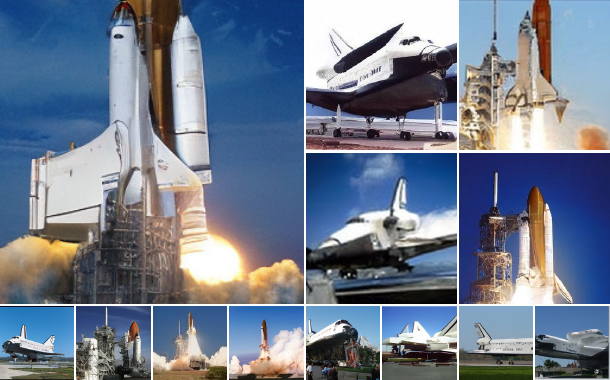}
\caption{Space shuttle (812)}
\end{subfigure}\hspace{6pt}
\begin{subfigure}{0.47\linewidth}
\includegraphics[width=\linewidth]{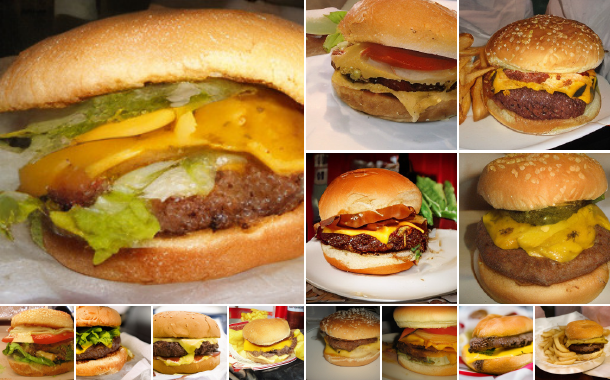}
\caption{Cheeseburger (933)}
\end{subfigure}
\caption{\textbf{Samples of NEPA-DiT-XL} on ImageNet \(256\times256\) (continued).}
\label{fig:samples-2}
\end{figure}

\begin{figure}[t]
\centering
\begin{subfigure}{0.47\linewidth}
\includegraphics[width=\linewidth]{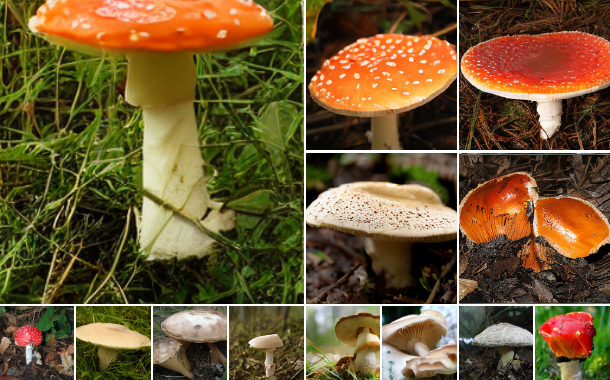}
\caption{Mushroom (947)}
\end{subfigure}\hspace{6pt}
\begin{subfigure}{0.47\linewidth}
\includegraphics[width=\linewidth]{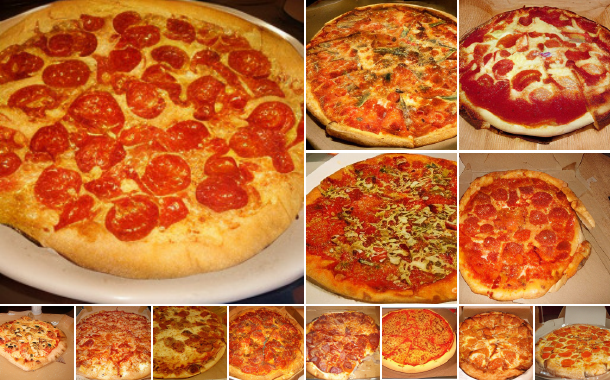}
\caption{Pizza (963)}
\end{subfigure}\\[4pt]
\begin{subfigure}{0.47\linewidth}
\includegraphics[width=\linewidth]{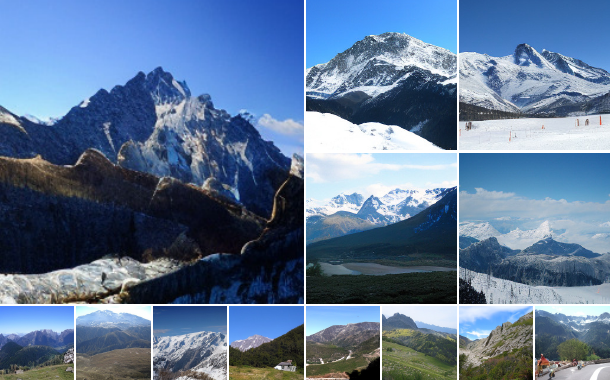}
\caption{Alp (970)}
\end{subfigure}\hspace{6pt}
\begin{subfigure}{0.47\linewidth}
\includegraphics[width=\linewidth]{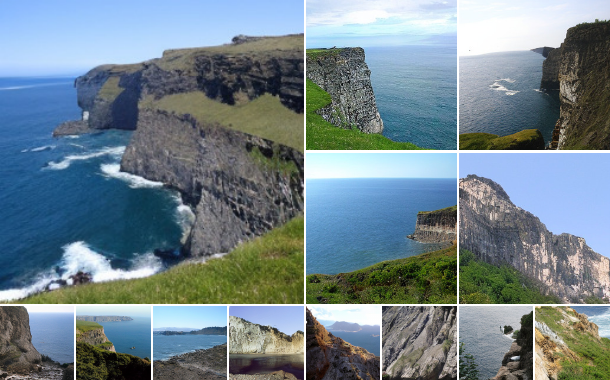}
\caption{Cliff (972)}
\end{subfigure}\\[4pt]
\begin{subfigure}{0.47\linewidth}
\includegraphics[width=\linewidth]{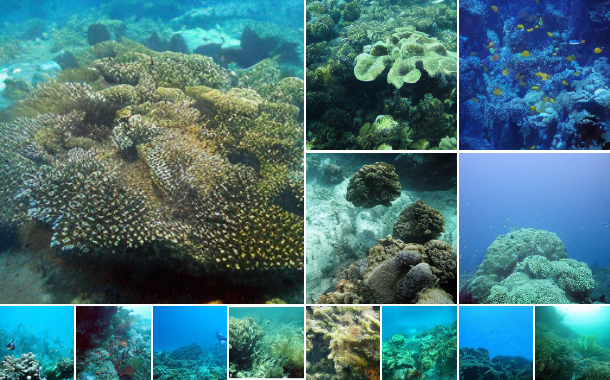}
\caption{Coral reef (973)}
\end{subfigure}\hspace{6pt}
\begin{subfigure}{0.47\linewidth}
\includegraphics[width=\linewidth]{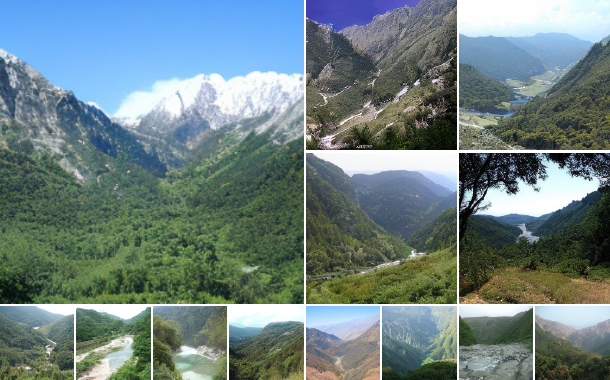}
\caption{Valley (979)}
\end{subfigure}\\[4pt]
\begin{subfigure}{0.47\linewidth}
\includegraphics[width=\linewidth]{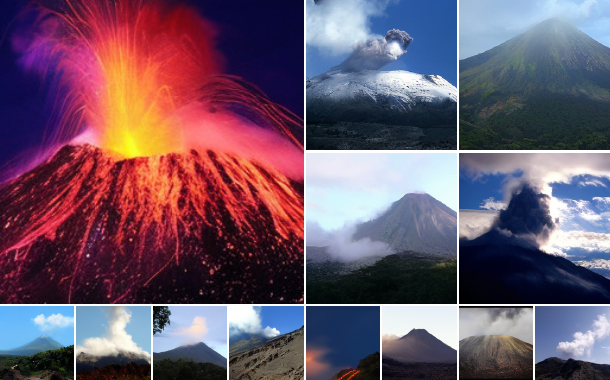}
\caption{Volcano (980)}
\end{subfigure}\hspace{6pt}
\begin{subfigure}{0.47\linewidth}
\includegraphics[width=\linewidth]{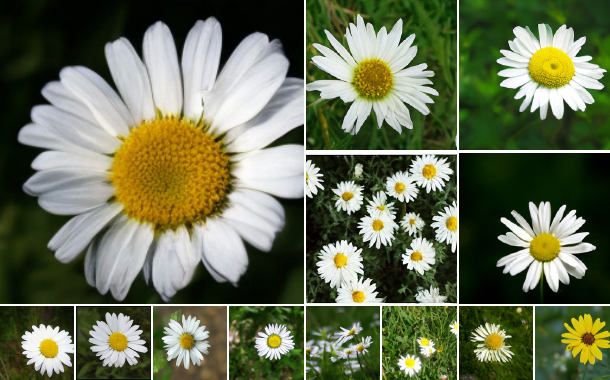}
\caption{Daisy (985)}
\end{subfigure}
\caption{\textbf{Samples of NEPA-DiT-XL} on ImageNet \(256\times256\) (continued).}
\label{fig:samples-3}
\end{figure}

Figures~\ref{fig:samples-1}--\ref{fig:samples-3} show more samples of NEPA-DiT-XL on ImageNet \(256\times256\), selected from 16 samples per class.

\end{document}